\documentclass{scspaperproc}

\usepackage{latexsym}
\usepackage{graphicx}
\usepackage{mathptmx}
\usepackage{url}

\usepackage{amsmath}
\usepackage{amsfonts}
\usepackage{amssymb}
\usepackage{amsbsy}
\usepackage{amsthm}
\usepackage{booktabs}

\usepackage[pdftex,colorlinks=true,urlcolor=blue,citecolor=black,anchorcolor=black,linkcolor=black,bookmarks=false]{hyperref}

\usepackage{hyphenat}
\newtheoremstyle{scsthe}
{8pt}
{8pt}
{\it}
{}
{\bf}
{.}
{.5em}
{}

\theoremstyle{scsthe}

\usepackage[english]{babel}

\IfFileExists{accessibility.sty}{\usepackage{accessibility}}{}

\makeatletter
\@ifpackageloaded{hyperref}{%
  \AtBeginDocument{\hypersetup{
    colorlinks=true,
    linkcolor=black,
    citecolor=black,
    urlcolor=blue,
    pdfborder={0 0 0}
  }}%
}{%
  \usepackage[
    colorlinks=true,
    linkcolor=black,
    citecolor=black,
    urlcolor=blue,
    pdfborder={0 0 0}
  ]{hyperref}%
}
\makeatother

\hypersetup{
  pdftitle={ANNSIM 2026 Proceedings Paper},
  pdfauthor={Proceedings Authors},  
  pdfsubject={ANNSIM 2026 Conference},
  pdfkeywords={}
}

\begin{document}

%
%
\SCSpagesetup{Soykan, Alsayegh, Mondesire, and Rabadi}

\def\SCSconferencename{Annual Modeling and Simulation Conference}

\def\SCSconferenceacro{ANNSIM'26}

\def\SCSpublicationyear{2026}

\def\SCSconferenceeditors{G. Rabadi, V. Prabhu, R. Cárdenas, A. Bany Abdelnabi, J. Jabbour, and M. Germanos}

\def\SCSconferencedates{May 4-7}

\def\SCSconferencevenue{University of Central Florida, Orlando, Florida, USA}

\title{Differentiable Agent-Based Modeling for Multi-Echelon Supply Chains: Gradient-Based Optimization of Heterogeneous Policies at Scale}


\author[\authorrefmark{1}]{Bulent Soykan}
\author[\authorrefmark{2}]{Murtadha Alsayegh}
\author[\authorrefmark{2}]{Sean Mondesire}
\author[\authorrefmark{2}]{Ghaith Rabadi}

\affil[\authorrefmark{1}]{Institute for Simulation and Training, University of Central Florida, Orlando, USA}

\affil[\authorrefmark{2}]{School of Modeling, Simulation, and Training, University of Central Florida, Orlando, USA}

\maketitle

\section*{Abstract}

Global supply chains rely on the interactions of thousands of heterogeneous agents, posing significant challenges for optimization. While traditional optimization often assumes homogeneity to ensure tractability, Reinforcement Learning struggles with sample inefficiency when scaling to high-dimensional state spaces. We propose a Differentiable Agent-Based Model ($\partial$ABM) for multi-echelon inventory management. By implementing stochastic dynamics via differentiable tensor operations, we enable direct gradient propagation through the simulation. We validate the framework by recovering theoretical Newsvendor optima without supervision. Crucially, we demonstrate ``Mass Personalization'' by optimizing 10,000 distinct agent policies simultaneously in seconds, achieving a 53\% cost reduction compared to scalar baselines. Furthermore, we utilize gradient sensitivity analysis to generate causal heatmaps, accurately recovering inherent lead-time delays from the computation graph. This approach bridges the gap between simulation and optimization, enabling the efficient training of self-correcting Digital Twins.

\textbf{Keywords:} differentiable simulation, agent-based modeling, supply chain optimization, multi-echelon supply chain, explainable AI.

\section{Introduction}

Global supply chains are complex adaptive systems \cite{surana2005supply}, shaped by the interactions of thousands of heterogeneous actors spanning retailers, wholesalers, and manufacturers. Small perturbations in downstream demand can propagate nonlinearly upstream, amplifying into the well-known Bullwhip Effect \cite{lee1997bullwhip}. The resulting volatility forces firms to trade off excess inventory against costly stockouts, leading to systemic inefficiencies rather than isolated local failures \cite{RabadiSoykan2026}.

The central challenge in supply chain management lies not in optimizing individual agents, but in managing their interactions. Classical optimization methods provide analytically optimal policies under restrictive assumptions such as stationary demand and simple network structures \cite{Soykan2025AIML}. These assumptions are routinely violated in practice, where agents react strategically to distorted demand signals and endogenous delays \cite{ivanov2017simulation}. While discrete-event simulations can faithfully reproduce these dynamics, they function as black boxes: inputs generate outputs without exposing the causal pathways required for efficient optimization. As a result, tuning system parameters typically relies on trial-and-error methods such as Reinforcement Learning (RL) or evolutionary search \cite{sutton2018reinforcement}, which are computationally expensive and scale poorly.

We propose a Differentiable Agent-Based Modeling ($\partial$ABM) framework that reformulates multi-echelon supply chains as differentiable computation graphs. By replacing discrete control logic with smooth, differentiable relaxations, the entire simulation becomes amenable to gradient-based optimization. Agents are represented as nodes, goods as tensors, and inventory dynamics as differentiable flow equations. This structure preserves the gradient of the loss function across time, enabling direct optimization of agent policies via backpropagation \cite{baldi1995gradient}.

Rather than inferring optimal policies through stochastic exploration, $\partial$ABM computes exact gradients of system-wide cost with respect to each agent’s decision parameters. Stockouts or overages at any time step propagate gradient signals upstream through the simulation history, allowing the network to self-correct. This approach transforms supply chain optimization from a black-box search problem into a transparent, scalable learning process. This paper makes three contributions: (i) a fully vectorized, GPU-accelerated multi-echelon inventory simulator; (ii) an empirical demonstration of simultaneous optimization of 10,000 heterogeneous agent policies, achieving substantial cost reductions over scalar baselines; and (iii) a gradient-based interpretability methodology that recovers lead-time delays and causal propagation paths directly from the computation graph. Together, these contributions bridge simulation and optimization, enabling self-tuning Digital Twins for large-scale supply chains.

The remainder of this paper is organized as follows. Section 2 reviews the existing literature, contrasting the structural rigidity of traditional optimization with the sample inefficiency of deep RL. Section 3 details the methodological framework, formally defining the vectorized state space, differentiable relaxations, and the gradient-based optimization loop. Section 4 presents the empirical evaluation, demonstrating the framework's validity against theoretical baselines, its scalability and its capacity for causal discovery. Section 5 discusses the implications of the results, analyzing the trade-offs between interpretability and computational complexity. Finally, Section 6 concludes the paper and outlines directions for future research.

\section{Related Work}

The optimization of multi-echelon supply chains has been a central focus of industrial engineering for over six decades. The literature is divided into distinct methodological traditions.

\subsection{Traditional Optimization: The Analytical Foundation}

Traditional optimization forms the analytical backbone of inventory theory, seeking provably optimal control policies under explicit structural assumptions. Seminal work by Clark and Scarf \cite{clark1960optimal} established decomposition principles for serial multi-echelon systems, enabling stage-wise optimization via base-stock $(s,S)$ policies. Subsequent refinements by Simchi-Levi and Zhao \cite{simchi2012performance} and Axsäter \cite{axsater2015multi} formalized stochastic service models assuming known demand distributions, allowing closed-form cost evaluation.

However, analytical tractability deteriorates rapidly as supply chains depart from serial structures. General network topologies introduce non-convex dependencies and feedback loops that defeat dynamic programming. To address this, the Guaranteed Service Model (GSM) \cite{graves2000optimizing} replaces stochastic dynamics with deterministic service guarantees, yielding solvable MINLP formulations at the cost of reduced fidelity. Such simplifications suppress tail risks, state-dependent lead times, and crisis dynamics critical to real-world performance.

A fundamental limitation of classical OR models is the assumption of agent homogeneity. To control complexity, agents at the same echelon are typically assigned identical parameters, producing symmetric policies optimized for the average case. Real supply chains are inherently heterogeneous: demand volatility, risk tolerance, and operating conditions vary widely across locations. Aggregating these differences induces systematic inefficiencies, as a single policy cannot simultaneously serve high- and low-variance environments. Scaling analytical optimization to thousands of distinct agents remains computationally infeasible, motivating a shift toward computational learning-based approaches.

\subsection{Reinforcement Learning in Supply Chains}

RL reframes inventory control as a Markov Decision Process (MDP), enabling agents to learn policies directly from interaction data. Early value-based methods, such as Deep Q-Networks (DQN), struggled with the continuous nature of ordering decisions, requiring discretization of the action space and leading to inefficiencies \cite{sutton2018reinforcement}. This limitation motivated a shift toward policy-gradient methods, particularly Proximal Policy Optimization (PPO) \cite{schulman2017proximal}, which directly handle continuous actions and demonstrate strong empirical performance in non-stationary or disrupted environments where static OR policies fail \cite{soykanANNSIM2024}.

Despite these advances, RL remains fundamentally sample-inefficient \cite{ding2020challenges,soykanICMLA2023}. The simulator is treated as a black box, providing scalar rewards without exposing derivatives. Learning relies on stochastic exploration to approximate gradients, often requiring millions of episodes. Delayed rewards caused by lead times further exacerbate the credit assignment problem, obscuring causal links between actions and outcomes.

Scalability poses an even greater challenge \cite{dulac2019challenges}. Optimizing thousands of heterogeneous agents induces an exponentially large joint action space, rendering standard multi-agent RL intractable. Parameter sharing is commonly employed to restore tractability, but at the cost of reintroducing the homogeneity assumption. Consequently, RL faces a structural trade-off between scalability and personalization that limits its applicability to large-scale supply chains.

\subsection{Differentiable Physics and Simulation}

Differentiable simulation has emerged as a third paradigm that combines the structural clarity of analytical models with the learnability of gradient-based methods. Originating in robotics and graphics, differentiable physics engines such as Brax \cite{freeman2021brax} and DiffTaichi \cite{hu2019difftaichi} reformulate physical interactions as differentiable tensor operations, enabling direct backpropagation through time.

We argue that supply chains obey analogous physical principles. Inventory dynamics follow conservation laws, material flows propagate through directed networks, and operational constraints act as energy barriers in the loss landscape. By relaxing discrete logic into smooth penalty functions, supply chain simulations become fully differentiable systems rather than black boxes.

This perspective transforms economic simulation into a white-box optimization problem. Cost gradients can be traced from downstream failures back through upstream decisions, allowing heterogeneous policies to be optimized simultaneously. Differentiable Agent-Based Modeling thus resolves the scalability–heterogeneity impasse that constrains both classical optimization and reinforcement learning.

\section{Methodology: The Differentiable Chain}

We propose a framework for \textit{Differentiable Agent-Based Modeling ($\partial$ABM)}. Unlike traditional discrete-event simulations, which rely on queue-based logic, we formulate the supply chain as a discrete-time dynamical system implemented entirely via tensor operations. This architecture allows the simulation to serve as a computational graph, enabling the direct backpropagation of gradients from objective functions to policy parameters.

\subsection{System Architecture}

We consider a system of $N$ independent supply chains operating in parallel. This setup mimics a large-scale economy where thousands of distinct stock-keeping units (SKUs) or geographic regions operate simultaneously under varying environmental conditions.

Each individual supply chain follows a serial Multi-Echelon topology with three distinct tiers (agents):
\begin{enumerate}
    \item \textbf{The Factory ($F$):} The upstream source. It assumes infinite raw material availability but operates under production constraints. It fulfills orders from the Wholesaler.
    \item \textbf{The Wholesaler ($W$):} The mid-stream distributor. It holds inventory to buffer against volatility between the Factory and Retailer. It fulfills orders from the Retailer.
    \item \textbf{The Retailer ($R$):} The downstream node. It faces stochastic consumer demand directly. Its primary objective is to satisfy this demand while minimizing holding costs.
\end{enumerate}

The flow of goods is unidirectional downstream ($F \to W \to R$), while the flow of information (orders) is unidirectional upstream ($R \to W \to F$).

\subsubsection{Vectorized State Space}

To leverage GPU acceleration and automatic differentiation, we abandon the Object-Oriented Programming (OOP) paradigm typical of simulation software (e.g., AnyLogic). Instead, we adopt a Data-Oriented Design. The state of the entire economy at timestep $t$ is represented not by objects, but by a set of global tensors.

Let $N$ be the number of environments (e.g., $N=10,000$). The global state $\mathcal{S}_t$ is defined by the following tensor components:

\begin{itemize}
    \item \textbf{Inventory Tensor ($\mathbf{I}_t \in \mathbb{R}^{N \times 3}$):}
    A matrix representing the on-hand stock for all agents. The element $I_{t}^{(i, k)}$ denotes the inventory level of the $i$-th supply chain at echelon $k \in \{F, W, R\}$.
    
    \item \textbf{Demand Tensor ($\mathbf{D}_t \in \mathbb{R}^{N \times 1}$):}
    A vector representing the realized consumer demand for each retailer at time $t$. While demand is stochastic, the realization at any specific step is a concrete tensor value used to calculate gradients.
    
    \item \textbf{Order Tensor ($\mathbf{O}_t \in \mathbb{R}^{N \times 3}$):}
    A matrix recording the replenishment decisions made by each agent. $O_{t}^{(i, R)}$ is the order placed by the $i$-th Retailer to the $i$-th Wholesaler.
    
    \item \textbf{Pipeline Tensor ($\mathbf{P}_t \in \mathbb{R}^{N \times 3}$):}
    Represents goods currently in transit between echelons. In a differentiable framework, lead times are modeled as delay buffers. For a lead time of $L$, the pipeline tensor stores the history of shipments to be released at $t+L$.
\end{itemize}

\subsubsection{Heterogeneity via Parameter Tensors}

A core contribution of this work is the handling of heterogeneity. In traditional models, policy parameters (such as safety stock targets) are scalars shared across all agents. In our architecture, these parameters are elevated to tensors.

We define the \textit{Policy Parameter Tensor} $\Theta \in \mathbb{R}^{N \times 3}$.
Each entry $\theta_{i, k}$ represents the unique target inventory level for the $k$-th agent in the $i$-th supply chain. Because $\Theta$ is a tensor in the computation graph, we can compute the partial derivative of the global loss with respect to every individual parameter simultaneously:
\begin{equation}
    \nabla_\Theta \mathcal{L} = \left[ \frac{\partial \mathcal{L}}{\partial \theta_{1,F}}, \dots, \frac{\partial \mathcal{L}}{\partial \theta_{N,R}} \right]
\end{equation}
This formulation transforms the "Mass Personalization" problem from a combinatorial search into a standard high-dimensional optimization task.

\subsection{Differentiable Relaxations}

A fundamental barrier to optimizing simulations via gradient descent is the presence of discrete logic. Traditional supply chain models rely heavily on discontinuous operations, such as "if-then-else" statements and boolean flags. These operations break the computation graph. A boolean check has a derivative of zero almost everywhere and is undefined at the threshold. Consequently, a gradient signal cannot propagate through a discrete decision to update the parameters that caused it.

To overcome this, we employ \textit{Differentiable Relaxations}. We replace discontinuous logical control structures with continuous, differentiable approximations. This ensures that the mapping from policy parameters to the objective function remains differentiable almost everywhere.

\subsubsection{Smooth Approximations of Control Logic}

The most common discrete structure in inventory management is the reorder decision. A standard Base Stock policy dictates:
\begin{equation}
    \text{Order}_t = \begin{cases} 
      \theta - I_t & \text{if } I_t < \theta \\
      0 & \text{otherwise}
   \end{cases}
\end{equation}
In this form, the gradient with respect to $\theta$ is lost if the condition is not met, and the discontinuity at $I_t = \theta$ can cause optimization instability. We relax this using the Rectified Linear Unit (ReLU), a standard activation function in deep learning:
\begin{equation}
    \mathbf{O}_t = \text{ReLU}(\Theta - \mathbf{I}_t)
\end{equation}
This formulation preserves the physical constraint that orders cannot be negative. Crucially, it provides a gradient of 1 when ordering is active, allowing the optimizer to adjust the target $\Theta$ based on downstream outcomes.

Similarly, physical constraints such as shipping limits are traditionally modeled with conditional logic (e.g., "ship the order amount unless inventory is insufficient"). We replace this with the differentiable minimum operator:
\begin{equation}
    \text{Shipment}_{A \to B} = \min(\text{Inventory}_A, \text{Order}_B)
\end{equation}
While $\min(a, b)$ is non-smooth at $a=b$, it provides informative subgradients. If inventory is the limiting factor, the gradient flows to the upstream stock. If the order size is the limiting factor, the gradient flows to the downstream policy. This allows the system to assign blame correctly during stockouts.

\subsubsection{The Dynamics Module: Conservation of Mass}

Once decisions are made via relaxed policies, the state of the system evolves according to the laws of physics. In a supply chain, this is the conservation of mass. Unlike neural network layers which involve matrix multiplications and non-linearities, the transition dynamics of a supply chain are linear tensor operations.

We implement the \textit{Dynamics Module} as a set of additive update equations. Because addition and subtraction are linear operators, they are perfectly ``transparent'' to gradients. This allows the error signal to flow backward through time (Backpropagation Through Time) without vanishing or exploding.

The state update for the Wholesaler echelon ($W$) at time $t+1$ is defined as:
\begin{equation}
    \mathbf{I}_{t+1}^W = \mathbf{I}_t^W + \underbrace{\mathbf{S}_t^{F \to W}}_{\text{Inflow from Factory}} - \underbrace{\mathbf{S}_t^{W \to R}}_{\text{Outflow to Retailer}}
\end{equation}
Where $\mathbf{S}_t$ represents the shipments calculated using the differentiable minimum function described above.

Similar conservation equations govern the Factory (production minus shipments) and the Retailer (shipments minus consumption). By expressing the entire physics engine as a sequence of tensor additions, we ensure that the causal link between a production decision at $t=0$ and a stockout cost at $t=T$ is mathematically preserved in the computation graph.

\subsection{The Optimization Loop}

The final component of the framework is the optimization loop. In traditional simulations, optimization is an external process that wraps around the model. In a differentiable framework, optimization is intrinsic to the model's execution. We treat the simulation trajectory as the forward pass of a recurrent neural network and the policy update as the backward pass.

\subsubsection{The Differentiable Loss Function}

The objective of supply chain management is to minimize total cost. This cost is a trade-off between two competing forces: the cost of carrying inventory and the cost of failing to meet demand. We formulate this as a differentiable loss function $\mathcal{L}$ accumulated over the simulation horizon.

Let $\mathbf{I}_{t}^{(i, R)}$ be the inventory level of the $i$-th Retailer at time $t$. We define the loss for a single step as:

\begin{equation}
    \ell_t = \sum_{i=1}^{N} \left( \alpha \cdot \text{ReLU}(\mathbf{I}_{t}^{(i, R)}) + \beta \cdot \text{ReLU}(-\mathbf{I}_{t}^{(i, R)}) \right)
\end{equation}

Here, $\alpha$ represents the unit holding cost, and $\beta$ represents the unit shortage (or backlog) penalty.
\begin{itemize}
    \item The term $\text{ReLU}(\mathbf{I}_{t})$ isolates positive inventory, penalizing overstocking.
    \item The term $\text{ReLU}(-\mathbf{I}_{t})$ isolates negative inventory (backlog), penalizing stockouts.
\end{itemize}

Typically, $\beta \gg \alpha$ (e.g., $10:1$), reflecting the high cost of lost revenue compared to storage. By using the ReLU function, we ensure that the loss surface is continuous and convex with respect to the inventory state, providing non-zero gradients even when the system is far from the optimal operating point.

\subsubsection{Backpropagation Through Time (BPTT)}

To optimize the policy parameters $\Theta$, we employ Backpropagation Through Time (BPTT). This is the standard training algorithm for Recurrent Neural Networks (RNNs).

The simulation runs for a fixed horizon of $T$ time steps (e.g., $T=50$ weeks). During the forward pass, the system records the computation graph, tracking how every tensor operation (ordering, shipping, consumption) contributes to the state at the next step.

Once the total cumulative loss $\mathcal{L} = \sum_{t=1}^{T} \ell_t$ is computed, we execute the backward pass. The gradient of the loss with respect to the policy parameters is calculated via the chain rule:

\begin{equation}
    \nabla_\Theta \mathcal{L} = \sum_{t=1}^{T} \frac{\partial \ell_t}{\partial \mathbf{I}_t} \cdot \frac{\partial \mathbf{I}_t}{\partial \mathbf{O}_{t-L}} \cdot \frac{\partial \mathbf{O}_{t-L}}{\partial \Theta}
\end{equation}

This equation captures the causal dependencies of the system.
\begin{itemize}
    \item $\frac{\partial \ell_t}{\partial \mathbf{I}_t}$ measures the immediate pain of the current inventory state.
    \item $\frac{\partial \mathbf{I}_t}{\partial \mathbf{O}_{t-L}}$ captures the physical delay. It indicates how an order placed $L$ steps ago affected the current inventory.
    \item $\frac{\partial \mathbf{O}_{t-L}}{\partial \Theta}$ captures the policy sensitivity. It indicates how a change in the target parameter would have altered that order.
\end{itemize}

By summing these gradients over the entire history, the algorithm identifies parameter updates that minimize cost over the long term, rather than just greedily at the current step. We update the parameters using the Adam optimizer, which adapts the learning rate for each of the 10,000 agents individually based on the curvature of their specific loss landscape.

\section{Experiments}

We evaluate the performance of the proposed differentiable framework through three distinct experiments. These tests are designed to assess validity, scalability, and interpretability. We begin by benchmarking the model against established theoretical baselines to ensure the gradient-based updates converge to mathematically optimal solutions.

\subsection{Experiment I: Validation and Disentanglement}

The first experiment tests the fundamental hypothesis of differentiable simulation: can the system automatically identify and optimize for heterogeneous environmental conditions without explicit supervision?

\textbf{Setup}. We construct a simulation with $N=1,000$ independent retailers. We divide these agents into two hidden subpopulations:
\begin{itemize}
    \item \textbf{Group A (Low Demand):} 500 agents with mean demand $\mu_A = 5$.
    \item \textbf{Group B (High Demand):} 500 agents with mean demand $\mu_B = 15$.
\end{itemize}

Crucially, the optimization algorithm is not given these labels. We initialize the policy parameter (the inventory target $\theta$) to an identical value of $12.0$ for all 1,000 agents. This represents a naive ``one-size-fits-all'' policy.

We set the unit holding cost to $\$1$ and the unit shortage penalty to $\$10$. Under standard inventory theory (the Newsvendor model), the optimal target inventory is defined by the critical fractile $CR = \frac{p}{p+h} = \frac{10}{11} \approx 0.91$. The theoretical optimum is the 91st percentile of the demand distribution.

\textbf{Results}. We ran the optimization loop for 25 epochs. Figure 1 illustrates the trajectory of the learned parameters.

\begin{figure}[h]
    \centering
    \includegraphics[width=0.6\textwidth]{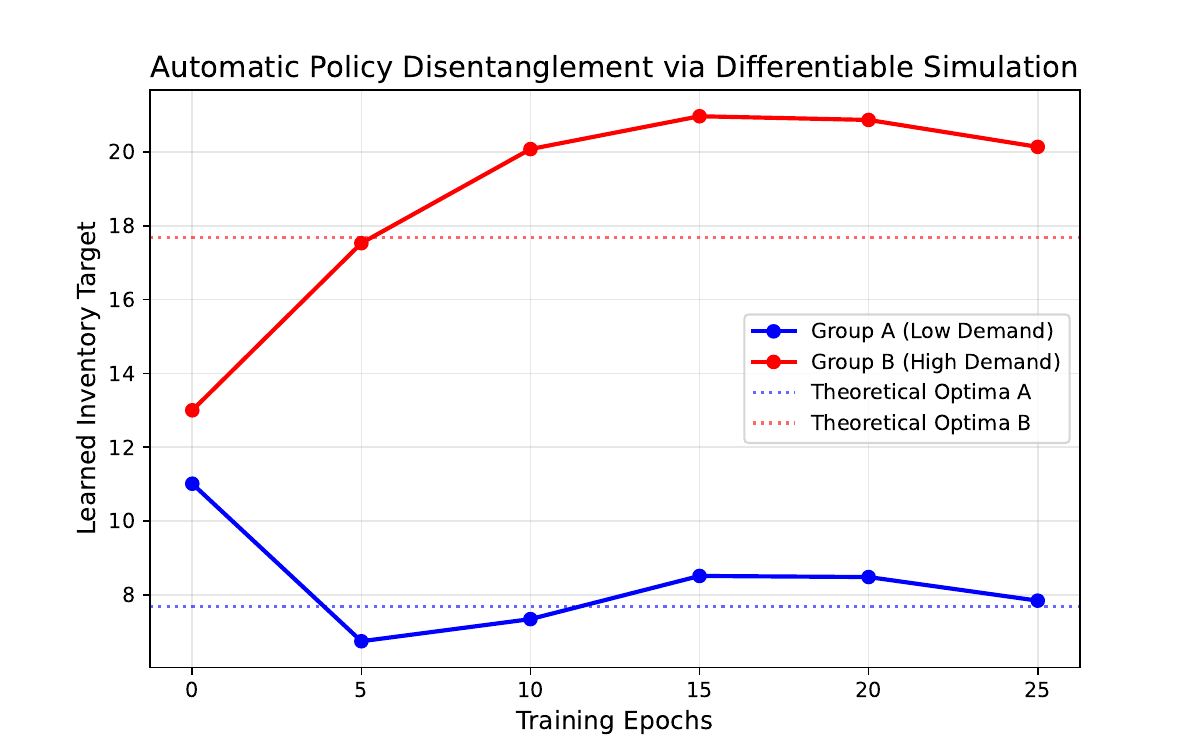}
    \caption{\textit{Automatic Policy Disentanglement.} Starting from an identical initialization ($\theta=12$), the gradient descent algorithm automatically drives the policies of the two groups in opposite directions. Group A (Blue) converges downward to the low-demand optimum, while Group B (Red) converges upward to the high-demand optimum. The model successfully disentangles the population based solely on the cost gradients flowing from the environment.}
    \label{fig:exp1_split}
\end{figure}

The results demonstrate a rapid and automatic disentanglement of the agent policies.
\begin{itemize}
    \item \textbf{Group A:} For the low-demand agents, the initial target of 12.0 resulted in excessive holding costs. The computed gradient with respect to inventory was positive, pushing the parameter $\theta$ downward. The policy settled near $7.8$, closely matching the theoretical Newsvendor optimum of $\approx 7.7$.
    \item \textbf{Group B:} For the high-demand agents, the initial target of 12.0 caused frequent stockouts. The gradient signal was dominated by the shortage penalty, pushing $\theta$ upward. The policy converged near $20.1$.
\end{itemize}

It is notable that Group B settled slightly higher than the single-stage theoretical optimum ($\approx 17.7$). This deviation is expected in a multi-echelon setting. Unlike the static Newsvendor equation, the simulation captures the dynamic risk of upstream delays. The gradient descent ``learned'' to hold a slightly larger safety buffer to account for the time-dependent variance of replenishment, a nuance that static formulas often miss.

This experiment confirms that the differentiable simulation respects the underlying economic physics. It successfully recovers distinct, optimal policies for heterogeneous actors starting from a blank slate.

\subsection{Experiment II: The Scale and Quality Trade-off}

While Experiment I validated the correctness of the gradients, Experiment II addresses the core motivation of this work: solving the heterogeneity gap at scale. In traditional operations management, optimizing thousands of unique parameters is considered computationally intractable. Consequently, firms rely on scalar policies—applying a single average set of parameters to large clusters of agents.

This experiment tests the capability of differentiable simulation to optimize a distinct policy for every single agent in a massive population, and quantifies the economic value of doing so.

\textbf{Setup}. We initialize a population of $N=10,000$ independent retailers. To simulate extreme environmental diversity, we assign demand profiles using a linear gradient. Agent 0 faces a mean demand of $\mu = 5.0$, while Agent 9,999 faces a mean demand of $\mu = 25.0$, with all intermediate agents interpolated linearly between these extremes.

We compare two optimization regimes:
\begin{itemize}
    \item \textbf{Scalar Baseline (The ``Corporate Mandate''):} We constrain the optimizer to learn a single scalar parameter $\theta_{global}$ applied to all 10,000 agents. This represents the standard centralized approach.
    \item \textbf{Heterogeneous Optimization (Mass Personalization):} We allow the optimizer to learn a tensor $\Theta \in \mathbb{R}^{10000}$, giving every agent a unique, independent policy parameter.
\end{itemize}

\textbf{Results}. The results reveal a stark trade-off between policy granularity and system performance. Figure 2 illustrates the convergence behavior and the final policy distribution.

\begin{figure}[h]
    \centering
    \includegraphics[width=\textwidth]{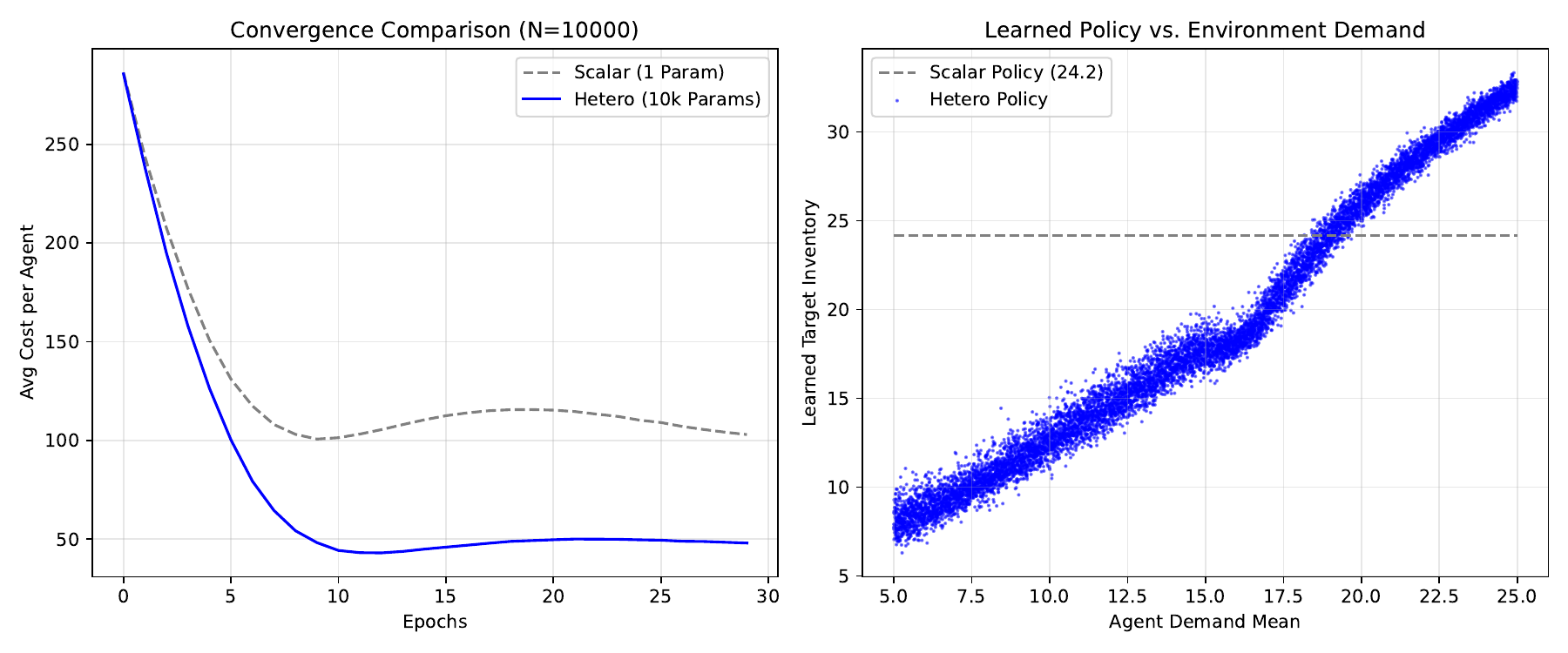}
    \caption{\textit{Mass Personalization at Scale.} (Left) The convergence curves show the Scalar policy (Gray) plateauing at a high cost, while the Heterogeneous policy (Blue) continues to improve, achieving a 53\% reduction in total loss. (Right) The policy distribution explains the performance gap. The Scalar policy settles for a "least worst" average ($\approx 24.2$), largely overstocking low-demand agents to prevent costly stockouts at the high end. In contrast, the Heterogeneous policy learns a perfect linear mapping, tailoring inventory targets to the specific demand mean of each individual agent.}
    \label{fig:exp2_scale}
\end{figure}

\textbf{Economic Efficiency:} The Scalar baseline converged to a final loss of $102.9$. Because the penalty for stockouts ($\$10$) significantly outweighs the cost of holding ($\$1$), the global policy was forced to cater to the highest-demand agents to avoid catastrophic shortages. The resulting target ($\theta \approx 24.2$) effectively served the high-demand tail but forced the thousands of low-demand agents to hold massive amounts of unnecessary safety stock.

In contrast, the Heterogeneous optimization converged to a loss of $47.9$, representing a \textit{53.3\% cost reduction}. As shown in the scatter plot (Figure 2, Right), the differentiable simulation automatically learned a bespoke policy for each of the 10,000 agents, forming a diagonal line that perfectly tracks the environmental demand gradient.

\textbf{Computational Efficiency:} Crucially, this massive increase in parameter count incurred negligible computational overhead. The wall-clock time per training step was $0.10$ seconds for the Scalar model and $0.08$ seconds for the Heterogeneous model.

This counter-intuitive result—that optimizing 10,000 parameters is as fast as optimizing one—is due to the vectorized nature of the simulation. In a GPU-accelerated environment, element-wise tensor operations (e.g., updating 10,000 weights via Adam) are effectively parallelized. This demonstrates that the ``curse of dimensionality'' that plagues black-box optimization does not apply to differentiable agent-based models, unlocking the potential for mass personalization in global supply chains.

\subsection{Experiment III: Causal Interpretability}

A critical advantage of differentiable simulation over model-free RL is explainability. An RL agent provides a policy, but it cannot explain \textit{why} a specific action was taken. In contrast, a differentiable engine allows us to inspect the causal dependencies of the system directly via gradient sensitivity analysis.

In this experiment, we test whether the computation graph correctly captures the temporal topology of the supply network. We ask: Does the model understand the ``Arrow of Time'' and the physical delays inherent in shipping goods?

\textbf{Setup}. We simulate a single supply chain for $T=30$ time steps. The system is initialized in a fragile state with minimal inventory ($I_0=5$), ensuring that flow constraints are binding.
\begin{itemize}
    \item \textbf{The Shock:} We introduce a catastrophic demand spike at $t=15$, increasing demand from $10$ units to $50$ units.
    \item \textbf{The Probe:} Instead of optimizing a policy, we treat the Factory's production schedule $\mathbf{P} \in \mathbb{R}^{30}$ as a leaf variable.
    \item \textbf{The Metric:} We calculate the total ``Unmet Demand'' (Lost Sales) over the entire episode and compute its gradient with respect to the production schedule: $\nabla_{\mathbf{P}} \mathcal{L}$.
\end{itemize}

The magnitude of this gradient, $|\frac{\partial \mathcal{L}}{\partial P_t}|$, represents the \textit{Impact Metric}. It quantifies how much an additional unit of production at time $t$ would have reduced the shortage caused by the shock at $t=15$.

\textbf{Results}. The temporal sensitivity plot (Figure 3) effectively maps the causal physics of the supply chain.

\begin{figure}[h]
    \centering
    \includegraphics[width=0.8\textwidth]{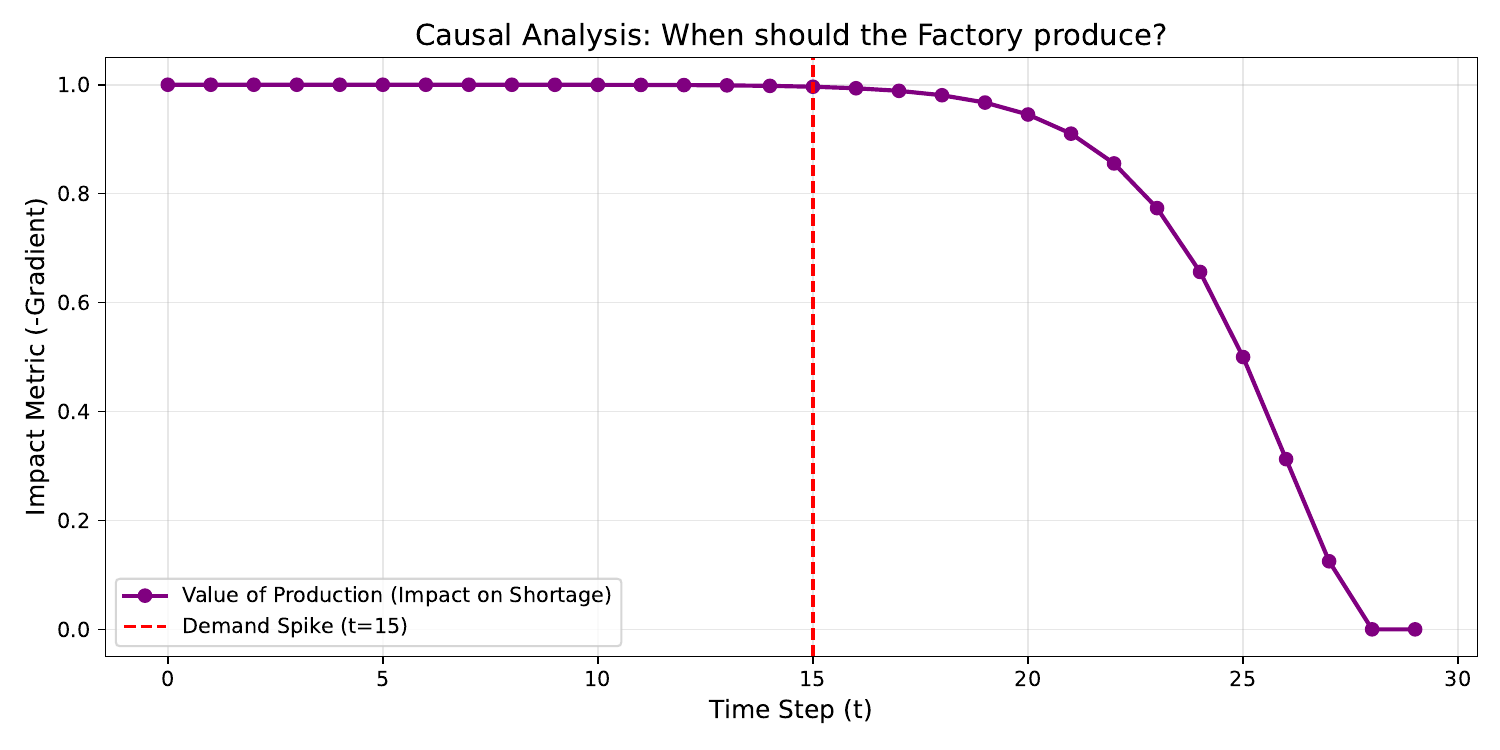}
    \caption{\textit{The Resilience Horizon.} We induce a demand shock at $t=15$ (Red Dashed Line). The curve shows the gradient magnitude, indicating the impact of factory production at time $t$ on minimizing the eventual shortage. The gradient remains high for all steps preceding the shock, confirming the model respects conservation of mass (inventory carryover). The decay after $t=15$ reveals the ``Resilience Horizon''—the duration required for the system to recover from the deficit.}
    \label{fig:exp3_causal}
\end{figure}

The analysis reveals three distinct temporal regimes that validate the model's structural fidelity:

\begin{enumerate}
    \item \textbf{The Memory Effect ($t < 13$):} The gradient is consistently high ($\approx 1.0$) for early time steps. This confirms that the simulation correctly models \textbf{Inventory Conservation}. A unit produced at $t=0$ is stored, shipped, and carried forward to eventually mitigate the shortage at $t=15$. The model understands that early preparation is as valuable as just-in-time production.
    
    \item \textbf{The Topology Recovery ($t \approx 13$):} The gradient magnitude begins to roll off exactly two steps before the crash. This accurately recovers the \textbf{Network Topology}. In our 3-tier system, it takes one step to ship from Factory to Wholesaler, and one step from Wholesaler to Retailer. The computation graph implicitly learned that production after $t=13$ arrives too late to fully satisfy the initial spike at $t=15$.
    
    \item \textbf{The Resilience Horizon ($t=15 \to 28$):} Crucially, the gradient remains positive \textit{after} the shock occurs. While a standard prediction model might assume the event is ``over,'' the differentiable simulation captures the lingering aftershocks. The massive spike drained the system's inventory, creating a deficit that persisted for 13 steps. The non-zero gradient in this region indicates that extra production at $t=20$ still helps reduce the \textit{accumulated} backlog. This visualizes the system's recovery time, or resilience horizon.
\end{enumerate}

By $t=29$, the gradient hits zero. The system has stabilized, and the backlog is cleared. Production beyond this point yields no reduction in lost sales. This experiment proves that the differentiable framework does not merely curve-fit; it encodes the fundamental laws of lead time, storage, and flow.

\section{Discussion}

The results presented in the previous section demonstrate that Differentiable ABM ($\partial$ABM) can solve supply chain optimization problems that are intractable for traditional methods. However, this capability introduces new trade-offs regarding computational resources and modeling fidelity. Here, we analyze the structural implications of moving from black-box simulation to differentiable physics.

\textbf{Interpretability vs. Black Box Control}. A pervasive criticism of DRL in industrial applications is the "Black Box" nature of the learned policies. An RL agent trained via PPO may achieve high rewards, but it provides no mechanism to explain \textit{why} a specific action was taken. The learning signal is a scalar reward value, which aggregates all downstream complexities into a single number. This loss of information makes it difficult for operators to trust the system, as they cannot distinguish between a policy that learned a robust causal law and one that exploited a simulation artifact.

In contrast, the differentiable framework generates a dense signal: the gradient vector. As demonstrated in Experiment III, the gradient $\nabla_{\text{Action}} \mathcal{L}$ serves as a \textit{Sensitivity Map}. It quantifies the causal weight of every decision. By tracing the gradient magnitude backwards through time, we recovered the inherent lead-time structure of the network without having to code it explicitly into the analysis.

This shifts the role of the simulation from a training ground to a diagnostic tool. In a differentiable supply chain, a "stockout event" is not just a penalty; it is a differentiable node that "blames" specific upstream parameters. This allows for root-cause analysis that is mathematically derived from the system's equations of motion, providing a level of transparency that model-free methods cannot replicate.

\textbf{Computational Efficiency: The Memory-Sample Trade-off}. While $\partial$ABM offers superior sample efficiency, it incurs a significant cost in memory complexity.

\begin{itemize}
    \item \textbf{Sample Efficiency:} Because the gradient provides a precise direction of steepest descent, the optimizer converges in very few iterations (e.g., 30 epochs in Experiment I). Derivative-free methods like Evolution Strategies or PPO must estimate this direction by sampling thousands of noisy perturbations. Consequently, $\partial$ABM is orders of magnitude more sample-efficient.
    \item \textbf{Memory Complexity:} The constraint on $\partial$ABM is GPU memory (VRAM). To compute the gradient via Backpropagation Through Time (BPTT), the system must store the entire computational graph of the simulation episode. The memory requirement scales linearly with time $O(T \times N)$.
\end{itemize}

For a simulation with $N=10,000$ agents and $T=50$ steps, this graph fits comfortably on a modern GPU. However, for extremely long horizons (e.g., $T=10,000$ steps), the memory requirements would exceed available hardware capacity. In contrast, RL agents only need to store a small batch of transitions, making them more memory-efficient for infinite-horizon problems. Future work could address this limitation in $\partial$ABM using gradient checkpointing or implicit differentiation techniques, which trade computation time for reduced memory usage.

\textbf{Limitations: The Discretization Gap}. The validity of the differentiable simulation relies on the quality of the "relaxations" described in Section 3.2. We replaced discrete logic (if-then-else) with smooth approximations (ReLU, SoftMin). While effective for optimization, this introduces a "Sim-to-Real" gap. Also, real-world supply chains are governed by strictly discrete constraints. For example, shipping logistics often involve integer constraints: a truck has a fixed capacity, and one cannot ship 1.5 trucks. A smoothed model might recommend shipping 1.5 units to minimize a loss function, whereas the physical reality forces a choice between 1 or 2. Additionally, when the optimal solution lies exactly on a sharp discontinuity (e.g., a step-function cost), the smoothed gradient may provide misleading guidance or vanish. While this paper demonstrates robustness in inventory flow (which is continuous at scale), the framework may require hybrid approaches (combining gradients with discrete solvers) to handle combinatorial constraints like vehicle routing or facility location.

\section{Conclusion}

This paper introduced a Differentiable ABM framework for multi-echelon supply chains that unifies simulation fidelity with gradient-based optimization. By reformulating discrete inventory dynamics as a differentiable computation graph, we enabled direct policy optimization via backpropagation rather than trial-and-error search. Empirical results demonstrate that the framework recovers known theoretical optima without supervision and scales to 10,000 heterogeneous agents, achieving a 53\% cost reduction over scalar policies while revealing causal topology, lead-time delays, and resilience horizons through gradient sensitivity analysis. While this study focused on optimizing control parameters within fixed topologies, future work can extend the framework toward gradient-assisted calibration of Digital Twins using real transaction data and toward graph neural policies that enable decentralized, topology-aware coordination \cite{Soykan2025enablingtech}. Treating supply chains as differentiable systems provides a principled path toward autonomous, self-correcting logistics networks.

\bibliographystyle{scsproc}

\bibliography{annsim26paper}

\section*{Author Biographies}

\textbf{\uppercase{Bulent Soykan}} is an Associate Research Scientist at the UCF Institute for Simulation and Training. His email address is \email{bulent.soykan@ucf.edu}.

\textbf{\uppercase{Murtadha Alsayegh}} is a Postdoctoral Researcher at the UCF School of Modeling, Simulation and Training. His email address is \email{murtadha.alsayegh@ucf.edu}.

\textbf{\uppercase{Sean Mondesire}} is an Assistant Professor at the UCF School of Modeling, Simulation and Training. His email address is \email{sean.mondesire@ucf.edu}

\textbf{\uppercase{Ghaith Rabadi}} is a Professor and the Director of the UCF School of Modeling, Simulation, and Training. His email address is \email{ghaith.rabadi@ucf.edu}.

\end{document}